\documentclass[letterpaper]{article} 
\usepackage[draft]{aaai25}  
\usepackage{times}  
\usepackage{helvet}  
\usepackage{courier}  
\usepackage[hyphens]{url}  
\usepackage{graphicx} 
\usepackage{natbib}  
\usepackage{caption} 
\usepackage{algorithm}
\usepackage{algorithmic}
\usepackage{amsmath}
\usepackage{amssymb}
\usepackage{svg}
\usepackage{multirow}
\usepackage{booktabs}
\usepackage{url}
\usepackage{newfloat}
\usepackage{listings}
\DeclareCaptionStyle{ruled}{labelfont=normalfont,labelsep=colon,strut=off} 
\floatstyle{ruled}
\newfloat{listing}{tb}{lst}{}
\floatname{listing}{Listing}
\title{SpotActor: Training-Free Layout-Controlled Consistent Image Generation}

\author{
    Jiahao Wang\textsuperscript{\rm 1},
    Caixia Yan\textsuperscript{\rm 1,}\thanks{Corresponding authors.},
    Weizhan Zhang\textsuperscript{\rm 1,}\footnotemark[1],
    Haonan Lin\textsuperscript{\rm 1},
    Mengmeng Wang\textsuperscript{\rm 2},\\
    Guang Dai\textsuperscript{\rm 3},
    Tieliang Gong\textsuperscript{\rm 1},
    Hao Sun\textsuperscript{\rm 4},
    Jingdong Wang\textsuperscript{\rm 5}
}
\affiliations{
    \textsuperscript{\rm 1}Xi’an Jiaotong University\\
    \textsuperscript{\rm 2}Zhejiang University of Technology\\
    \textsuperscript{\rm 3}SGIT AI Lab\\
    \textsuperscript{\rm 4}China Telecom Artificial Intelligence Technology Co.Ltd\\
    \textsuperscript{\rm 5}Baidu Inc\\

    \{uguisu,linhaonan\}@stu.xjtu.edu.cn, \{yancaixia,zhangwzh,gongtl\}@xjtu.edu.cn, wangmengmeng@zjut.edu.cn, gdai@gmail.com, sunh10@chinatelecom.cn, wangjingdong@outlook.com
}
\begin{document}

\maketitle

\begin{abstract}
Text-to-image diffusion models significantly enhance the efficiency of artistic creation with high-fidelity image generation. However, in typical application scenarios like comic book production, they can neither place each subject into its expected spot nor maintain the consistent appearance of each subject across images. For these issues, we pioneer a novel task, \textit{Layout-to-Consistent-Image} (L2CI) generation, which produces consistent and compositional images in accordance with the given layout conditions and text prompts. To accomplish this challenging task, we present a new formalization of \textit{dual energy guidance} with optimization in a dual semantic-latent space and thus propose a training-free pipeline, \textbf{SpotActor}, which features a layout-conditioned backward update stage and a consistent forward sampling stage. In the backward stage, we innovate a nuanced layout energy function to mimic the attention activations with a sigmoid-like objective. While in the forward stage, we design \textit{Regional Interconnection Self-Attention} (RISA) and \textit{Semantic Fusion Cross-Attention} (SFCA) mechanisms that allow mutual interactions across images. To evaluate the performance, we present \textbf{ActorBench}, a specified benchmark with hundreds of reasonable prompt-box pairs stemming from object detection datasets. Comprehensive experiments are conducted to demonstrate the effectiveness of our method. The results prove that SpotActor fulfills the expectations of this task and showcases the potential for practical applications with superior layout alignment, subject consistency, prompt conformity and background diversity.
\end{abstract}

\section{Introduction}
Diffusion probabilistic models \cite{ddpm,ldm,dpm1} have achieved notable success in the realm of image generation. Within this domain, text-to-image (T2I) diffusion models \cite{CFG,sdxl} enable artists to generate high-quality images with descriptions of their desired subjects. Thus, their applicability extends to numerous practical contexts for their substantial contributions to artistic productivity. Despite the success, however, their performance in some application scenarios still exhibits aspects in need of further refinement. For instance, in real-world creation scenarios like comic book drawing, a natural process involves conceptualizing the appearance of a specific character, designing the visual layout of each scene, and then illustrating a series of images of the character. This process reflects two essential skills of professionals: the ability to preserve appearance consistency of the same character, and the capacity to render image content in alignment with the pre-defined layout\textemdash both of which are lacking in standard diffusion models.

\begin{figure}[t]
\centering
\includegraphics[width=0.46\textwidth]{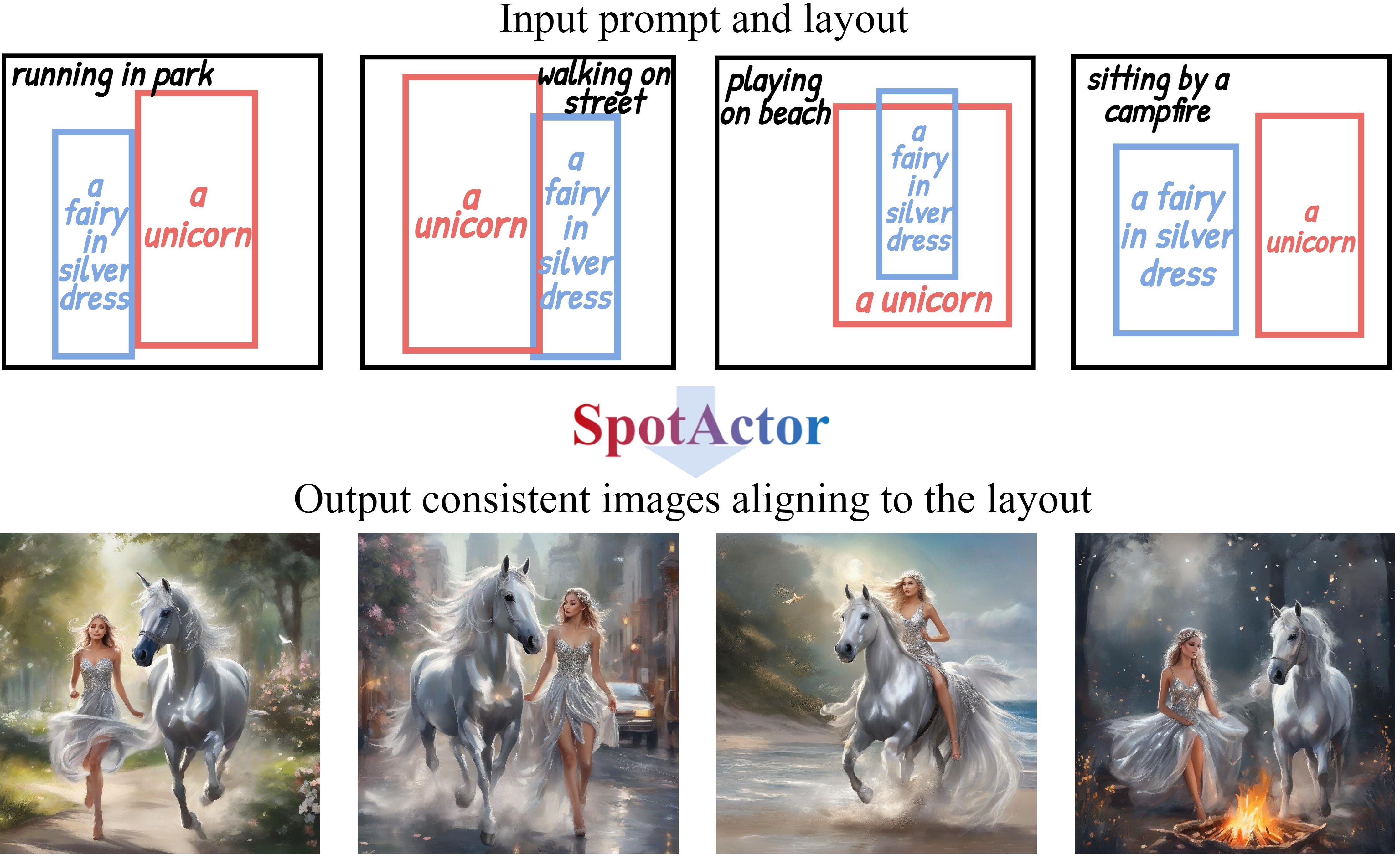}
\caption{Given bounding boxes and text prompts of subject and plot descriptions, our method generates high-quality images where subjects align to the layout as well as share a consistent appearance.}
\label{fig:teasor}
\end{figure}

Since the emergence of diffusion models, the challenges of subject consistency and layout controllability have continuously been two separate topics of ongoing research interest. For subject consistency, TheChosenOne \cite{thechosenone} first introduces the task of \textit{consistent subject generation}, which aims to generate consistent images of the same subject solely driven by prompts, and differentiate it from other analogous tasks. It proposes a tuning-based approach to accomplish this task yet its iterative backbone tuning process results in high computational expenses and may degrade the image quality. Later, OneActor \cite{oneactor} achieves a $4\times$ faster tuning speed without compromising the image quality via intricate cluster-conditioned guidance. More recently, training-free methods \cite{consistory,storydiffusion} are proposed to bypass the tuning process by enhancing the backbone with handicraft modules activated during the inference process. For the layout controllability, prevailing methods \cite{selfguidance,freecontrol,layoutguidance} manage to achieve the \textit{layout-to-image generation} task via energy guidance in a training-free manner. To specify, they regulate the latent codes through the backward propagation driven by customized energy functions and thus align the visual elements to their expected positions. Nevertheless, in the context of the aforementioned creation scenario, no existing work embarks on addressing both challenges simultaneously. Besides, existing works in either task pay limited attention to the role of the semantic space in T2I models.

For these issues, we pioneer a novel task, \textit{layout-to-consistent-image} (L2CI) generation. As shown in Fig.~\ref{fig:teasor}, given a series of expected bounding boxes, the corresponding subject descriptions (e.g. \textit{fairy}\&\textit{unicorn}) and the plot descriptions, this task aims to generate a series of images where subjects share the consistent appearance as well as situate perfectly in the given boxes. To accomplish this challenging task, we propose \textbf{SpotActor}, the first L2CI generation pipeline in a training-free manner. We start from the insight that the semantic space and spatial latent space of diffusion models are inherently entangled and share certain properties \cite{photomaker,oneactor}. Hence, we consider the semantic and latent space as a whole dual space and present a new formalization of \textit{dual energy guidance}, which defines an update trajectory in the dual space. The formalization splits the pipeline into two stages: a layout-conditioned backward update stage and a consistent forward sampling stage. In the backward stage, we design a sigmoid-like objective based on in-depth analysis of the network activations to regulate the attention distributions. The objective later drives a backward update of the latent codes and the semantic embeddings to search for an optimal alignment with the pre-defined boxes. Subsequently in the forward stage, we enhance the ordinary backbone with \textit{Regional Interconnection Self-Attention} (RISA) and \textit{Semantic Fusion Cross-Attention} (SFCA) mechanisms in order to allow inter-image level spatial-spatial interactions and spatial-semantic interactions according to the layout conditions, respectively. To evaluate the performance in this task, we present \textbf{ActorBench}, the first L2CI benchmark including hundreds of prompt-box pairs and a set of evaluation metrics. We creatively utilize real-world object detection datasets to construct the prompt-box pairs that comply with objective principles. Comprehensive experiments verify our motivation and demonstrate the effectiveness of our method. The balanced layout alignment, subject consistency, prompt conformity and background diversity confirm that SpotActor fulfills the expectations of this task.

To summarize, our main contributions are as follows:
\begin{itemize}
    \item We pioneer the layout-to-consistent-image generation task that aims to maintain a consistent appearance of subjects as well as align them to the given layout.
    \item We consider the semantic and latent space as a whole dual space and formalize a novel dual energy guidance to jointly optimize in the semantic-latent space.
    \item We propose SpotActor pipeline to address the L2CI task, which features a backward update based on the nuanced layout energy and a forward sampling enhanced by two intricate attention mechanisms.
    \item We present the first L2CI generation benchmark, ActorBench, and conduct comprehensive experiments to evaluate the effectiveness of our method.
\end{itemize}

\section{Related Work}
\subsection{Consistent Subject Generation}
This task is first proposed by TheChosenOne \cite{thechosenone} and focuses on generating images of the same subject based solely on descriptive prompts. The pioneer work presents a repetitive process of generating, clustering and tuning to regulate the generation distribution into a cohesive cluster. Yet the laborious backbone tuning takes 20 minutes to function and may harm the inner capacity of the backbone. Later, OneActor \cite{oneactor} proposes a cluster guidance paradigm, supersedes the backbone tuning with a projector optimization and reduces the required time to 5 minutes. More recently, training-free methods \cite{consistory,storydiffusion} are introduced to eliminate the tuning process by designing new self-attention mechanisms that directly function during the inference. Nonetheless, the training-free methods hardly pay attention to the spatial-semantic interaction of the diffusion process, and no attempts have been made to incorporate the layout controllability. Thus, we accomplish a novel layout-controlled consistent subject generation task by leveraging latent-semantic optimization.

\subsection{Layout-to-Image Generation}
As diffusion models prevail, numerous works manage to harness the diffusion backbone to generate images aligning with the given layout. Early tuning-based methods fine-tune the backbone with handicraft modules \cite{gligen} or specific token embeddings \cite{reco} to inject layout conditions, while training-free methods \cite{ediff,ldm,dense} achieve the goal by manipulating the attention procedure. Recently, a branch of training-free methods \cite{zeroshot,boxdiff,selfguidance,layoutguidance} is gaining prominence for its elegant energy guidance. To specify, they design a backward propagation based on the layout loss, which optimizes the latent codes to align with the given layout. However, optimizing solely in the latent space restricts the search range as the other half, the semantic space, is continuously neglected. Hence, we formalize a new dual energy guidance approach to jointly optimize the latent codes and semantic embeddings, unleashing the full potential of the diffusion model.

\section{Preliminaries}
\begin{figure*}[t]
\centering
\includegraphics[width=0.95\textwidth]{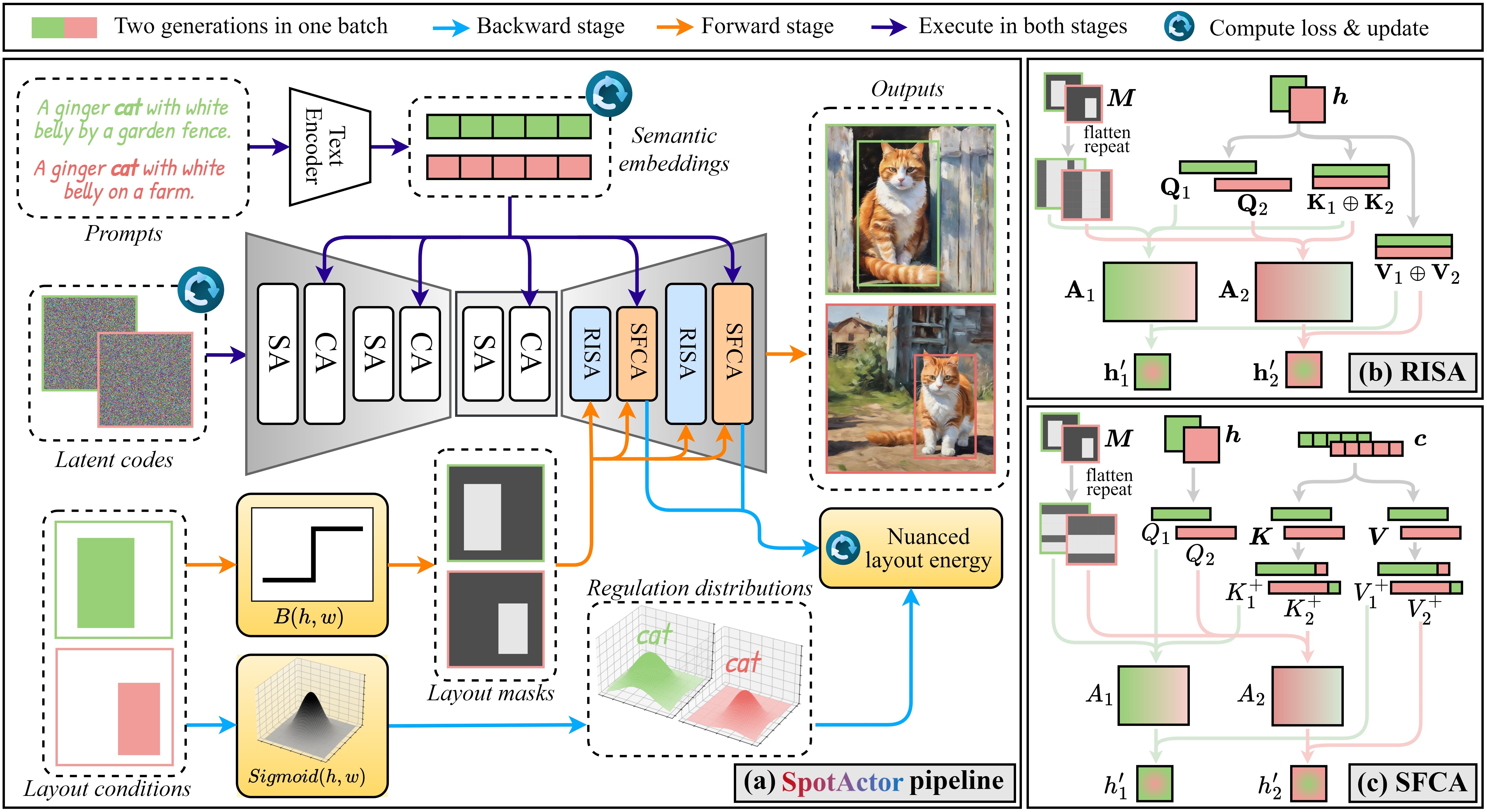}
\caption{The overall architecture of SpotActor. (a) Our method consists of two stages at each sample step in a dual energy guidance manner. The backward stage optimizes the latent codes and semantic embeddings with the nuanced layout energy based on the sigmoid-like objective. Subsequently, the forward sampling is enhanced by two intricate attention mechanisms: (b) RISA and (c) SFCA.}
\label{fig:framework}
\end{figure*}
Before introducing our method, we first provide a brief review of diffusion models. From the score-based perspective \cite{SBDM1,SBDM2}, \textit{diffusion models} \cite{diffusion,ddpm,ldm} essentially manage to estimate a score function of the latent distribution of real images, $\nabla_{\boldsymbol{z}_t}\log p(\boldsymbol{z}_t)$, where $\boldsymbol{z}_t$ is contaminated data from a predetermined, time-dependent noise addition process. A denoising network $\boldsymbol{\theta}$ is trained to estimate the score at each step: $\hat{\boldsymbol{\epsilon}}_t=\boldsymbol{\epsilon}_{\boldsymbol{\theta}}(\boldsymbol{z}_t,t,\boldsymbol{c})\approx-\sigma_t\nabla_{\boldsymbol{z}_t}\log p(\boldsymbol{z}_t)$, where $\sigma_t$ are pre-defined constants, $\boldsymbol{c}$ is the semantic embeddings of the given prompt. Then during generation, taking DDPM \cite{ddpm} as an example, high-quality images are sampled from random noise by iteratively predicting $\boldsymbol{z}_{t-1}$ from $\boldsymbol{z}_t$ based on the score in a step-by-step manner:
\begin{equation} \label{sample-step}
    \boldsymbol{z}_{t-1}=\frac1{\sqrt{1-\beta_t}}(\boldsymbol{z}_t+\beta_t\nabla_{\boldsymbol{z}_t}\log p(\boldsymbol{z}_t))+\sqrt{\beta_t}\boldsymbol{\boldsymbol{\epsilon}},
\end{equation}
where $\beta_t$ is a set of pre-defined constants and $\boldsymbol{\boldsymbol{\epsilon}}\sim\mathcal{N}(\mathbf{0},\mathbf{I})$. To incorporate more flexible control, \textit{energy guidance} \cite{energy1} suggests that any energy function $e(\boldsymbol{z}_t,t,\boldsymbol{c})$ can be leveraged like a score to update the latent codes for different purposes:
\begin{equation} \label{latent-backward}
\boldsymbol{z}_t\xleftarrow{}\boldsymbol{z}_t-v\sigma_t\nabla_{\boldsymbol{z}_t}e(\boldsymbol{z}_t,t,\boldsymbol{c}),
\end{equation}
where $v$ is the energy guidance scale.


Inside the denoising network, $\boldsymbol{\theta}$, commonly implemented as a U-Net \cite{unet}, self-attention layers project features of latent codes, $\boldsymbol{h}$, into queries, key and value through projection matrices $\mathbf{W}$: $\mathbf{Q}^\textit{sa}=\boldsymbol{h}\cdot \mathbf{W}_\mathbf{Q}^\textit{sa}$, $\mathbf{K}^\textit{sa}=\boldsymbol{h}\cdot \mathbf{W}_\mathbf{K}^\textit{sa}$, $\mathbf{V}^\textit{sa}=\boldsymbol{h}\cdot \mathbf{W}_\mathbf{V}^\textit{sa}$. While cross-attention layers project $\boldsymbol{h}$ into queries and project semantic embeddings into keys and values: $\mathbf{Q}^\textit{ca}=\boldsymbol{h}\cdot \mathbf{W}_\mathbf{Q}^\textit{ca}$, $\mathbf{K}^\textit{ca}=\boldsymbol{c}\cdot \mathbf{W}_\mathbf{K}^\textit{ca}$, $\mathbf{V}^\textit{ca}=\boldsymbol{c}\cdot \mathbf{W}_\mathbf{V}^\textit{ca}$. The outputs, $\boldsymbol{h}^\prime$, are then calculated via standard attention mechanism: $\mathbf{A}=\mathrm{softmax}\left(\mathbf{Q\cdot K}^\top/\sqrt{d_k}\right)$, $\boldsymbol{h}^\prime=\mathbf{A}\cdot \mathbf{V}$, where $d_k$ is the feature dimension of $\mathbf{W}_\mathbf Q$ and $\mathbf{W}_\mathbf K$.

\section{Method}
\subsection{Overview}
In this task, users input a batch of $N$ prompt embeddings, $\{\boldsymbol c_i\}_{i=1}^N$, with the token embeddings of the central subject, $\{\boldsymbol c_i^\textit{sub}\}_{i=1}^N$, and bounding boxes, $\{\boldsymbol b_i=(h^\textit{min}_i,w^\textit{min}_i,h^\textit{max}_i,w^\textit{max}_i)\}_{i=1}^N$. Our goal is to generate $N$ consistent images of the subject with respect to the layout boxes. For this purpose, we propose a L2CI generation pipeline, SpotActor, as shown in Fig.~\ref{fig:framework}. To elaborate, we first consider the latent and semantic space as a whole and formalize a new \textit{dual energy guidance} approach which consists of two stages at each generation step. The backward stage manages to place the subject in the desired location, in which we update the latent codes and semantic embeddings with nuanced layout energy based on in-depth activation analysis. Subsequently, the forward stage contributes to the consistent appearance of the subject, in which we enhance the U-Net sampling with \textit{Regional Interconnection Self-Attention} (RISA) and \textit{Semantic Fusion Cross-Attention} (SFCA) mechanisms in Figs.~\ref{fig:framework}(b) and \ref{fig:framework}(c). Note that though we illustrate with single subject generation for simplicity, our pipeline can be seamlessly extended to multiple subject generation.

\begin{figure}[t]
\centering
\includegraphics[width=0.45\textwidth]{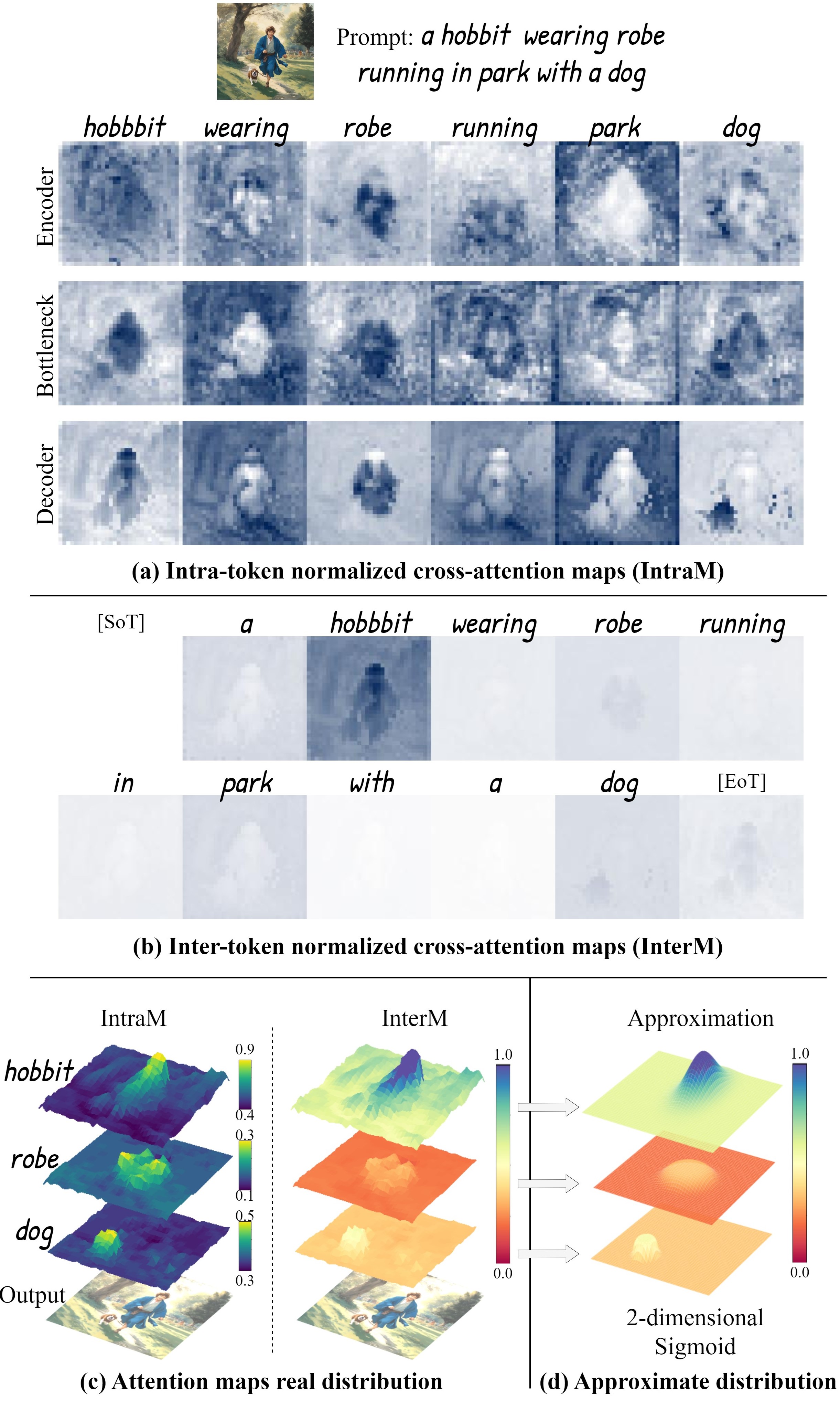}
\caption{Illustration of the attention analysis. (a) IntraM is the attention map normalized within each token, while (b) InterM is normalized across all the tokens. We further visualize (c) 3D distributions of attention maps and propose (d) sigmoid-like approximate distributions.}
\label{fig:analysis}
\end{figure}
\subsection{Formalization of Dual Energy Guidance}
The standard energy guidance consists of a backward update process $p_b$ and a forward sampling process $p_f$ to transform $\boldsymbol{z}_t$ to $\boldsymbol{z}_{t-1}$ at each step, which can be denoted as:
\begin{equation} \label{old-process}
    p(\boldsymbol{z}_{t-1}\mid \boldsymbol{z}_t)=p_f(\boldsymbol{z}_{t-1}\mid \boldsymbol{z}^*_t)\cdot p_b(\boldsymbol{z}^*_t\mid \boldsymbol{z}_t),
\end{equation}
where $\boldsymbol{z}^*_t$ is the optimized latent codes. In the backward stage, Eq.~(\ref{latent-backward}) with a control energy function is used to update the latent code until an optimal $\boldsymbol{z}^*_t$ that minimizes the control energy is found. Subsequently, the forward sampling in Eq.~(\ref{sample-step}) is performed to finish this step. However, the standard approach determines a sampling trajectory solely in the latent space and neglects the role of semantic condition. As proven in previous works \cite{photomaker,oneactor}, the latent space and the semantic space are inherently entangled together and ought to be regarded as a whole. Thus, we propose to reform Eq.~(\ref{old-process}) into:
\begin{equation} \label{new-process}
    \begin{aligned}
        p(\boldsymbol{z}_{t-1},\boldsymbol{c}_{t-1}\mid \boldsymbol{z}_t,\boldsymbol{c}_t)&=\\
        p_f(\boldsymbol{z}_{t-1},\boldsymbol{c}_{t-1}\mid &\boldsymbol{z}^*_t,\boldsymbol{c}^*_t)\cdot p_b(\boldsymbol{z}^*_{t},\boldsymbol{c}^*_t\mid \boldsymbol{z}_t,\boldsymbol{c}_t),
    \end{aligned}
\end{equation}
where $p_b$ and $p_f$ are layout-conditioned backward update and consistent forward sampling, which will be detailed in the following subsections.

\subsection{Layout-Conditioned Backward Update} \label{sec:backward}
\subsubsection{Composition Property of Cross-Attention.}
In the pursuit of designing an effective energy function for layout control without image quality degradation, we initiate our approach by analyzing the composition property of the cross-attention component. We conduct a standard generation process on SDXL \cite{sdxl} and collect the cross-attention maps $\mathbf{A}^\textit{ca}$ to explore the interaction between spatial pixels and semantic tokens. Fig.~\ref{fig:analysis}(a) presents the intra-token normalized maps (IntraM) from U-Net encoder layers, bottleneck layers, and decoder layers. It demonstrates that the activation of spatial pixels by semantic tokens corresponds to the composition of the final image, which echoes previous works \cite{ptp,pnp}. Beyond this established conclusion, we further observe that the correspondence intensifies with increasing network depth from encoder to decoder layers. Meanwhile, we normalize IntraM to the same scale to obtain the inter-token normalized maps (InterM) in Fig.~\ref{fig:analysis}(b) and thus reveal that different semantic tokens do not activate the spatial pixels equally, but exhibit different levels. These varying degrees of semantic-spatial interactions have a nuanced impact on the final image quality. We consider this discovery to be highly significant, yet it has been hardly utilized in existing works. 
\subsubsection{Nuanced Layout Energy Function.}The analysis above inspires us to transform the layout conditions into precise target distributions to regulate the generated subjects. To this end, we proceed with a detailed distribution analysis concentrating on three nouns in Fig.~\ref{fig:analysis}(c). We can observe that the activation of each noun exhibits a peak shape aligning with its spatial location at a certain range level. For example, the activation of the \textit{dog} token maintains high values in the center part of the \textit{dog} spatial area. When approaching the edge part, it sharply declines to lower values. To mimic the intra-token peak distribution, we extend Sigmoid function to 2-dimensional:
\begin{equation} \label{sigmoid}
    \mathrm{Sigmoid}(x,y)=\frac{1}{1+e^{-s\cdot(1-\frac{(x-\mu_1)^2}{\sigma_1}+\frac{(y-\mu_2)^2}{\sigma_2})}},
\end{equation}
where $(\mu_1,\mu_2)$ marks the center; $\sigma_1$ and $\sigma_2$
establish the margin and $s$ is a shape control factor. Yet for the inter-token distributions, directly regulating the range levels may degrade the image quality. For this issue, we employ spatial normalization to allow the model to allocate activation levels spontaneously. Therefore, the whole backward update can be articulated as below. Given a bounding box $\boldsymbol b=(h^\textit{min},w^\textit{min},h^\textit{max},w^\textit{max})$ and a subject token embedding $\boldsymbol c^{sub}$, we first define the target distribution in Eq.~(\ref{sigmoid}):
\begin{equation}
    \mu_1=\frac{h^\textit{min}+h^\textit{max}}{2}, \mu_2=\frac{w^\textit{min}+w^\textit{max}}{2},
\end{equation}
\begin{equation}
    \sigma_1=\frac{(h^\textit{max}-h^\textit{min})^2}{4}, \sigma_2=\frac{(w^\textit{max}-w^\textit{min})^2}{4}.
\end{equation}
We execute the forward sampling of U-Net to collect the cross-attention maps $\mathbf{A}^\textit{ca}\in\mathbb{R}^{K\times S}$ of the subject token from the U-Net decoder, averaged across different layers. $K$ is the number of attention heads and $S=H \times W$ is the total number of flattened pixels, where $H$ and $W$ are the height and width, respectively. We then reshape it and perform min-max normalization along the spatial dimensions to obtain $\tilde{\mathbf{A}}^\textit{ca}\in\mathbb{R}^{K\times H\times W}$. The energy function can then be calculated by:
\begin{equation}
    e=\frac{1}{KHW}\sum_k\sum_h\sum_w\left( \tilde{\mathbf{A}}^\textit{ca}_\textit{khw}-\mathrm{Sigmoid}(\frac{h}{H},\frac{w}{W})\right)^2,
\end{equation}
where $k$, $h$, $w$ are dimension indices. From the dual energy guidance perspective of Eq.~(\ref{new-process}), we simultaneously update the semantic embedding in the backward stage besides Eq.~(\ref{latent-backward}):
\begin{equation}
    \boldsymbol{c}_t\xleftarrow{}\boldsymbol{c}_t-w\sigma_t\nabla_{\boldsymbol{c}_t}e(\boldsymbol{z}_t,t,\boldsymbol{c}),
\end{equation}
until a batch of optimal $(\boldsymbol{z}^*_t,\boldsymbol{c}^*_t)$ are found. Note that even though the semantic update doesn't necessarily need $\sigma_t$, we add it as a dynamic step-wise weight.
\subsection{Consistent Forward Sampling}
For the semantic space, we define the forward sampling as $\boldsymbol{c}_{t-1}=\boldsymbol{c}^*_t$. While for the latent forward sampling, we enhance the ordinary U-Net with two attention mechanisms to maintain consistent appearance of the subject.
\subsubsection{Regional Interconnection Self-Attention.} The self-attention mechanism in ordinary U-Net enables the spatial pixels from one image to interact with each other, contributing to the final image with a consistent style and content. We desire to broaden the scope of this mechanism to an inter-image level with respect to the layout conditions, which gives rise to RISA. As illustrated in Fig.~\ref{fig:framework}(b), given binary layout masks $\mathbf M_{\textit{i}}\in \mathbb{R}^{H\times W}$ transformed from $\boldsymbol b_i$ and features of latent codes $\boldsymbol h_i$, we first flatten the masks and expand to obtain $\mathbf{M}^\textit{sa}_i\in\mathbb{R}^{K\times S\times S}$. Then we concatenate keys and values, respectively and utilize layout masks to precisely control the interconnection region:
\begin{equation}
        \mathbf{K}^\textit{sa+}=[\mathbf{K}^\textit{sa}_1\oplus \mathbf{K}^\textit{sa}_2\oplus\ldots\oplus \mathbf{K}^\textit{sa}_\textit{N}],
\end{equation}
\begin{equation}
        \mathbf{V}^\textit{sa+}=[\mathbf{V}^\textit{sa}_1\oplus \mathbf{V}^\textit{sa}_2\oplus\ldots\oplus \mathbf{V}^\textit{sa}_\textit{N}],
\end{equation}
\begin{equation}
        \mathbf{M}^\textit{sa+}_i=[\mathbf{M}^\textit{sa}_1\ldots \mathbf{M}^\textit{sa}_{i-1}\oplus \mathbf I\oplus \mathbf{M}^\textit{sa}_{i+1}\oplus\ldots \oplus \mathbf{M}^\textit{sa}_\textit{N}],
\end{equation}
\begin{equation}
    \boldsymbol h^{\textit{sa}}_i=\mathrm{Softmax}(\mathbf{Q}^\textit{sa}_i\cdot \mathbf{K}^{\textit{sa+}\top}/\sqrt{d_k}+\log \mathbf{M}^\textit{sa+}_i)\cdot \mathbf{V}^\textit{sa+},
\end{equation}
where $\mathbf I$ is matrix of ones, $\oplus$ indicates matrix concatenation and the superscript $^+$ represents the enlarged matrix.

\subsubsection{Semantic Fusion Cross-Attention.} The role of semantic space, as we highlight throughout our work, has been continuously overlooked in consistent generation works \cite{consistory,storydiffusion}. Hence, we design SFCA to enable each image to interact with all the semantic conditions within the batch. As shown in Fig.~\ref{fig:framework}(c), given $\mathbf M_{\textit{i}}$, $\boldsymbol h_i$ and the semantic embedding $\boldsymbol c_i$, $\mathbf{M}^\textit{ca}_i\in\mathbb{R}^{K\times S\times 1}$ is obtained by flattening, transposing and expanding. We then locate the corresponding $\mathbf K_i^\textit{sub}$ and $\mathbf V_i^\textit{sub}$ of the subject token and cross-concatenate them for a fused interaction within the layout region:
\begin{equation}
    \mathbf{K}^\textit{ca+}_i=[\mathbf{K}^\textit{sub}_1\oplus \ldots \mathbf{K}^\textit{sub}_{i-1}\oplus \mathbf{K}^\textit{ca}_i\oplus \mathbf{K}^\textit{sub}_{i+1}\oplus\ldots\oplus \mathbf{K}^\textit{sub}_\textit{N}],
\end{equation}
\begin{equation}
    \mathbf{V}^\textit{ca+}_i=[\mathbf{V}^\textit{sub}_1\oplus \ldots \mathbf{V}^\textit{sub}_{i-1}\oplus \mathbf{V}^\textit{ca}_i\oplus \mathbf{V}^\textit{sub}_{i+1}\oplus\ldots\oplus \mathbf{V}^\textit{sub}_\textit{N}],
\end{equation}
\begin{equation}
        \mathbf{M}^\textit{ca+}_i=[\mathbf{M}^\textit{ca}_1\ldots \mathbf{M}^\textit{ca}_{i-1}\oplus \mathbf I\oplus \mathbf{M}^\textit{ca}_{i+1}\oplus\ldots \oplus \mathbf{M}^\textit{ca}_\textit{N}],
\end{equation}
\begin{equation}
    \boldsymbol h^{\textit{ca}}_i=\mathrm{Softmax}(\mathbf{Q}^\textit{ca}_i\cdot \mathbf{K}^{\textit{ca+}\top}_i/\sqrt{d_k}+\log \mathbf{M}^\textit{ca+}_i)\cdot \mathbf{V}^\textit{ca+}_i.
\end{equation}

\section{Experiment}
\begin{figure*}[t]
\centering
\includegraphics[width=0.94\textwidth]{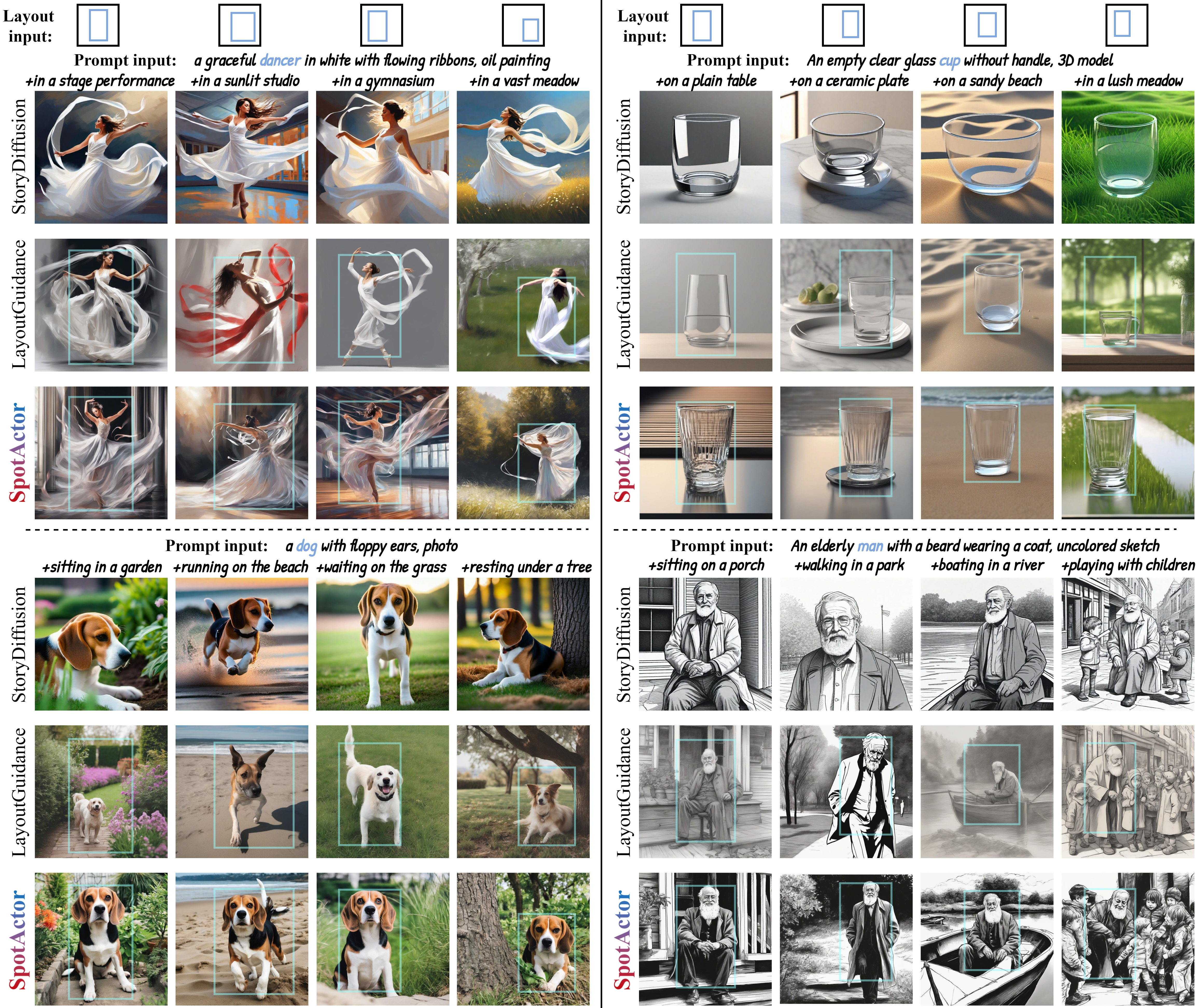}
\caption{The qualitative comparison between baselines and our SpotActor. Our method shows superior layout controllability compared to LayoutGuidance
and exhibits better subject consistency compared to StoryDiffusion. The central subjects are marked in blue and the given boxes are outlined in blue lines.}
\label{fig:comparison}
\end{figure*}
\begin{figure*}[t]
\centering
\includegraphics[width=0.9\textwidth]{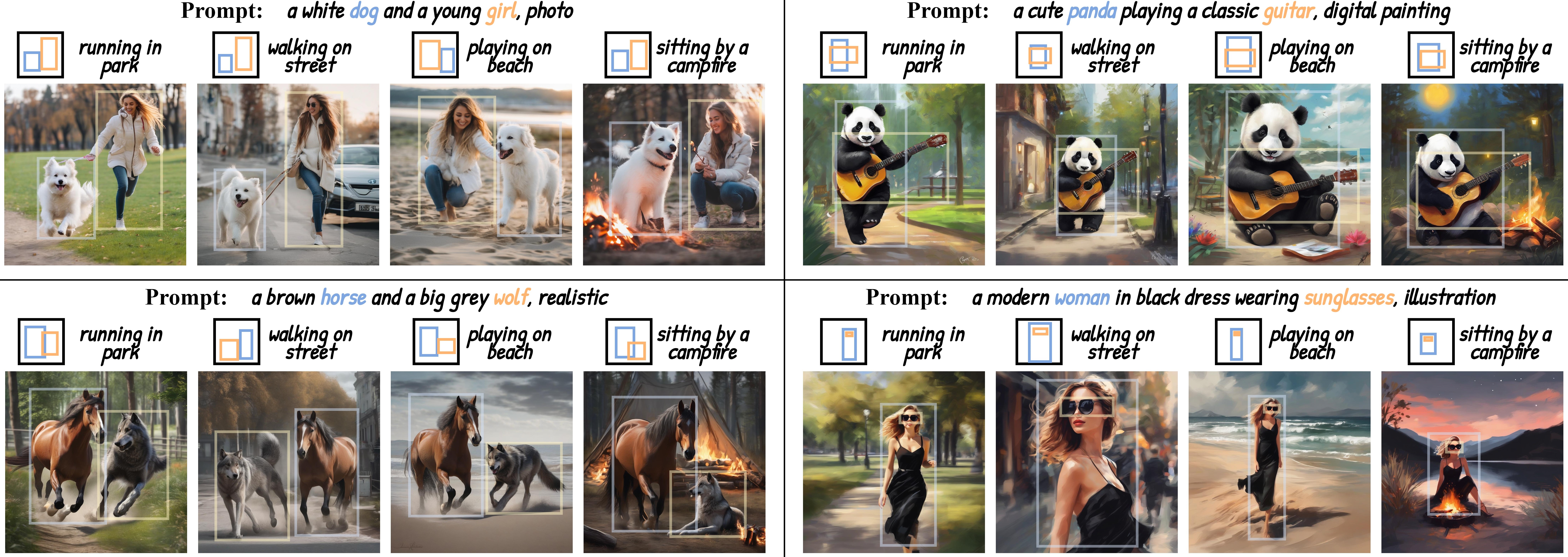}
\caption{Illustration of double subject generation by SpotActor. Our method maintains excellent performance when handling multiple subjects. Different central subjects are marked in different colors.}
\label{fig:multi}
\end{figure*}
\subsection{Actor-Bench}
To provide a fair and objective measurement of this novel task, we present ActorBench, the first layout-to-consistent-image generation benchmark. It includes 100 single-subject sets and 100 double-subject sets. Every set consists of four prompt-box pairs of the same central subject(s).
\subsubsection{Prompts\&Boxes.} In the endeavor for prompts and boxes that comply with objective principles, we direct our focus on COCO2017 \cite{coco}, a detection dataset of real-world photos. We utilize the train set annotations to obtain naturally associated subject-box pairs. We first perform data cleaning to remove boxes that are excessively small or positioned too close to the edge. Subjects that are inherently uniform in appearance (e.g. \textit{apple}) or unlikely to be personalized (e.g. \textit{plane}) are also removed. Then we collect four boxes of the same subject as a single-subject set and four double-boxes of the same double-subjects that exist in the same image as a double-subject set. All the subjects are divided into three main types: \textit{human}, \textit{animal}, and \textit{object} and we instruct ChatGPT \cite{chatgpt} to randomly convert some subjects to other more creative subjects of the same type (e.g. \textit{dog}$\xrightarrow{}$\textit{dragon}) and retain the corresponding boxes. Finally, we instruct ChatGPT to generate four formatted prompts for every set of subject(s): [\textit{appearance}]+[\textit{action}]+[\textit{background}]+[\textit{style}]. Note that \textit{appearance} and \textit{style} remain the same within one set and \textit{action} is only applicable for \textit{human} and \textit{animal}.

\subsubsection{Metrics.}
For evaluation, we perform subject-driven segmentation using Grounded-SAM \cite{grounded-sam} to obtain the subject boxes and separate the foregrounds (fg) and backgrounds (bg) of generated images. DINO \cite{dino}, CLIP \cite{clip} and LPIPS \cite{lpips} are utilized to extract visual embeddings. We introduce 4 dimensions of metrics: (1) layout alignment: we report the mean IoU (mIoU) between the detected boxes and the given boxes; (2) subject consistency: we calculate the cosine similarity among the visual embeddings of foregrounds to obtain DINO-fg and CLIP-fg. LPIPS-fg is also computed; (3) prompt conformity: we report the CLIP-T-Score \cite{clipscore} among the whole images; (4) background diversity: we calculate the scores of backgrounds to obtain DINO-bg, CLIP-bg and LPIPS-bg.

\subsection{Baselines}
To comprehensively evaluate the performance of SpotActor, we establish two training-free state-of-the-art models as baselines: consistent subject generation pipeline, StoryDiffusion \cite{storydiffusion} and layout control pipeline, LayoutGuidance. All the baselines are implemented on SDXL \cite{sdxl}. Besides, we construct 3 ablation models: our model excluding backward update (w.o. BU), excluding RISA (w.o. RISA) or excluding SFCA (w.o. SFCA). More experimental results and implementation details are presented in the Appendix.
\subsection{Results}
\vspace{-0.1cm}
\subsubsection{Quantitative Evaluation}
\begin{table*}[]
\centering
\setlength{\tabcolsep}{1.1mm}{
\begin{tabular}{c|c|ccc|c|ccc}
\hline
    &\textit{Layout}                      & \multicolumn{3}{c|}{}                                                                   &\textit{Prompt}                      & \multicolumn{3}{c}{}                                                                    \\
     &\textit{alignment}                   & \multicolumn{3}{c|}{\multirow{-2}{*}{\textit{Subject consistency}}}                              & \textit{conformity}                  & \multicolumn{3}{c}{\multirow{-2}{*}{\textit{Background diversity}}}                              \\ \cline{2-9} 
\multirow{-3}{*}{Method}                                & mIoU(↑)                     & DINO-fg(↑)                  & CLIP-fg(↑)                  & LPIPS-fg(↓)                 & CLIP-T(↑)             & DINO-bg(↓)                  & CLIP-bg(↓)                  & LPIPS-bg(↑)                 \\ \hline
SDXL                                  & 32.1                        & 53.6                        & 54.9                        & 43.6                        & \textbf{65.4}               & {\underline{30.9}}                  & {\underline{38.9}}                  & \textbf{60.8}               \\
StoryDiffusion                        & 29.5                        & {\underline{75.2}}                  & {\underline{78.1}}                  & \textbf{33.1}               & 63.5                        & 38.6                        & 46.8                        & 57.3                        \\
LayoutGuidance                        & {\underline{53.7}}                  & 53.7                        & 55.3                        & 40.8                        & 54.3                        & \textbf{29.6}               & \textbf{35.7}               & {\underline{60.1}}                  \\ \hline
{\color[HTML]{9B9B9B} Ours w.o. BU}   & {\color[HTML]{9B9B9B} 30.8} & {\color[HTML]{9B9B9B} 80.1} & {\color[HTML]{9B9B9B} 79.5} & {\color[HTML]{9B9B9B} 34.5} & {\color[HTML]{9B9B9B} 64.8} & {\color[HTML]{9B9B9B} 31.2} & {\color[HTML]{9B9B9B} 37.6} & {\color[HTML]{9B9B9B} 60.4} \\
{\color[HTML]{9B9B9B} Ours w.o. RISA} & {\color[HTML]{9B9B9B} 67.4} & {\color[HTML]{9B9B9B} 60.4} & {\color[HTML]{9B9B9B} 68.6} & {\color[HTML]{9B9B9B} 39.7} & {\color[HTML]{9B9B9B} 63.4} & {\color[HTML]{9B9B9B} 36.6} & {\color[HTML]{9B9B9B} 42.8} & {\color[HTML]{9B9B9B} 57.5} \\
{\color[HTML]{9B9B9B} Ours w.o. SFCA} & {\color[HTML]{9B9B9B} 67.3} & {\color[HTML]{9B9B9B} 74.8} & {\color[HTML]{9B9B9B} 72.2} & {\color[HTML]{9B9B9B} 35.6} & {\color[HTML]{9B9B9B} 64.1} & {\color[HTML]{9B9B9B} 37.1} & {\color[HTML]{9B9B9B} 46.2} & {\color[HTML]{9B9B9B} 57.1} \\
Ours (full)                                 & \textbf{67.1}               & \textbf{78.6}               & \textbf{79.7}               & {\underline{34.9}}                  & {\underline{63.6}}                  & 37.8                        & 49.5                        & 56.8                        \\ \hline
\end{tabular}
}
\caption{The quantitative results of baselines, ablation models, and our SpotActor. All the values are represented in percentage form. The results of ablation models are marked in grey. The best and second-best results are denoted in bold and underlined.}
\label{tab:quan}
\end{table*}

We illustrate the single subject generation result of baselines and our method in Fig.~\ref{fig:comparison}. As shown, StoryDiffusion exhibits competent consistency of subject appearance among images yet fails to be controlled by layout. With energy guidance, LayoutGuidance is able to place the subjects according to the given bounding boxes, but its inadequate approximation of the activation distribution leads to imperfectly aligned subjects (e.g. \textit{cup}, \textit{dog}). By contrast, on the one hand, our SpotActor demonstrates superior layout controllability. Benefiting from the nuanced and smooth sigmoid-based energy function, the subjects seamlessly stick to the given edges and fill the whole boxes. On the other hand, our pipeline maintains subject consistency decently with the proposed intricate attention mechanisms. As illustrated in Fig.~\ref{fig:multi}, our SpotActor naturally facilitates the multiple objects generation and continues to perform effectively in this scenario. The results prove that our method effectively fulfills the expectations of the L2CI generation.

\subsubsection{Quantitative Evaluation}
In Tab. \ref{tab:quan}, we display the results of the quantitative metrics between our method and baselines, which consist of four evaluation dimensions. For layout alignment, our method achieves a remarkable 67.1\% on mIoU and surpasses the LayoutGuidance by a large margin, which demonstrates our superior layout controllability. Meanwhile, in the dimension of subject consistency, our method scores the best at 78.6\% and 79.7\% on DINO-fg and CLIP-fg, showing the excellent capacity to maintain a consistent appearance of subjects. For prompt conformity, our method is second only to the original model with a narrow 1.8\% margin, which proves that our method reserves the great prompt controllability of the original model. Furthermore, due to the inherent bias, consistent generation pipelines inevitably compromise background diversity compared to SDXL, and our method exhibits competitive performance with StoryDiffusion. To summarize, our model displays strong and balanced performance in the 4-dimensional quantitative evaluation.

\subsubsection{Ablation Study}
\begin{figure}[t]
\centering
\includegraphics[width=0.43\textwidth]{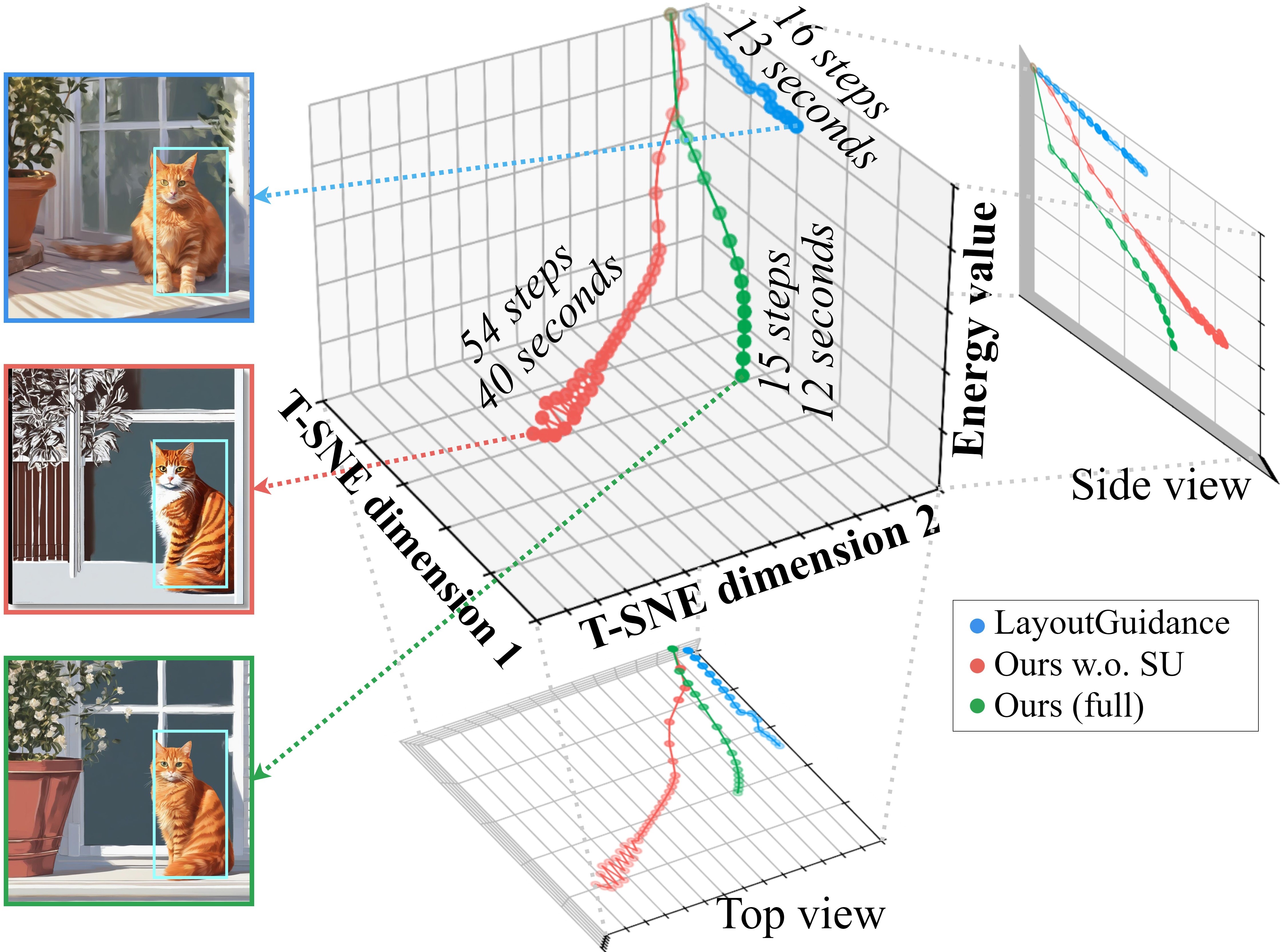}
\caption{The optimization trajectories of different guidance strategies. The XY-plane represents the dual space after T-SNE and the Z-axis corresponds to the normalized energy.}
\label{fig:ablation}
\end{figure}

To verify the effectiveness of each of our designs, we carry out a quantitative evaluation of the ablation models, as shown in Tab. \ref{tab:quan}. It demonstrates that the backward update (BU) benefits the model with an outstanding mIoU boost while the RISA and SFCA each contribute to the significant improvement of foreground scores. Besides, the backward and forward designs exhibit an ideal combined performance without interfering with each other. To evaluate the validity of dual energy guidance, which is the core of our work, we collect the latent codes, semantic embeddings and energy values in each update iteration of LayoutGuidance, our model excluding semantic update (ours w.o. SU) and our full model. We perform T-SNE to the combined latent codes and embeddings to represent the dual space with two dimensions. Thus, we visualize the optimization trajectories of the dual space with the energy value in Fig.~\ref{fig:ablation}. Since the energy definition of LayoutGuidance is different from ours, we normalize each sequence of energy values respectively. As illustrated, with dual energy guidance, our method rapidly converges to the optimum with improved layout alignment without image quality degradation.

\section{Conclusion}
This paper pioneers a novel training-free pipeline, SpotActor, for layout-to-consistent-image generation task. Considering the latent and semantic as a cohesive unit, we propose a new formalization of dual energy guidance including two stages. To perfectly align the subject to the given layout in the backward stage, we design a nuanced layout energy based on in-depth analysis. Later in the forward stage, we enhance the backbone with intricate attention mechanisms to strength the latent-semantic interactions, contributing to the consistent appearance of the generated subject. We further present a specialized benchmark, ActorBench, for evaluation. Comprehensive experiments highlight the effectiveness of our method with superior layout alignment, subject consistency as well as generation efficiency.

\nocite{ti,dreambooth,story1,story2,spatext}
\bibliography{aaai25}

\newpage
\appendix
\begin{huge}
    \noindent\textbf{Appendix}
\end{huge}
\section{Experiment Details}
\subsection{Implement Details}
All experiments are conducted on a single NVIDIA A800 80GB GPU. We implement our method on StableDiffusionXL (SDXL). All images are generated in 30 denoising steps and the inference guidance scale is set to be 5.0. We utilize layout-conditioned backward update to the first 3 steps and consistent forward sampling to the first 20 steps. During the backward update, we collect the attention maps from all decoder layers whose spatial dimension is 1024 to calculate the energy, The shape control factor of the sigmoid-like function is set to $s=10$. We set the latent energy guidance scale to $v=300$ and the semantic energy guidance scale to $w=0.9$. We undate with a convergence criterion with the energy of each subject lower than $k_\textit{thres}\times e_\textit{start}$, where $k_\textit{thres}$ is set to 60\% and $e_\textit{start}$ is the initial energy value. During the forward sampling, we employ the RISA and SFCA to all the decoder attention layers.
\subsection{Metric Evaluation}
We carry out experiments on our proposed benchmark, ActorBench. For each set of prompt-box pairs, we change random seeds to obtain 5 sets of generated images. All metrics are calculated by averaging the results.
\subsection{Package License}
All packages we utilize in this paper are credited as follows:
\begin{itemize}
    \item SDXL implementation at: \par \url{https://huggingface.co/stabilityai/stable-diffusion-xl-base-1.0}.
    \item StoryDiffusion implementation at: \par
    \url{https://github.com/HVision-NKU/StoryDiffusion}
    \item CLIP and DINO implementation at: \par \url{https://github.com/huggingface/transformers}.
    \item LPIPS implementation at: \par \url{https://github.com/richzhang/PerceptualSimilarity}.
    \item Grounded-SAM implementation at: \par
    \url{https://github.com/IDEA-Research/Grounded-Segment-Anything}
\end{itemize}

\newpage
\section{More qualitative Illustration}
\begin{figure}[h]
\setlength{\abovecaptionskip}{0.1cm}
\setlength{\belowcaptionskip}{-0.5cm}
\centering
\includegraphics[width=0.45\textwidth]{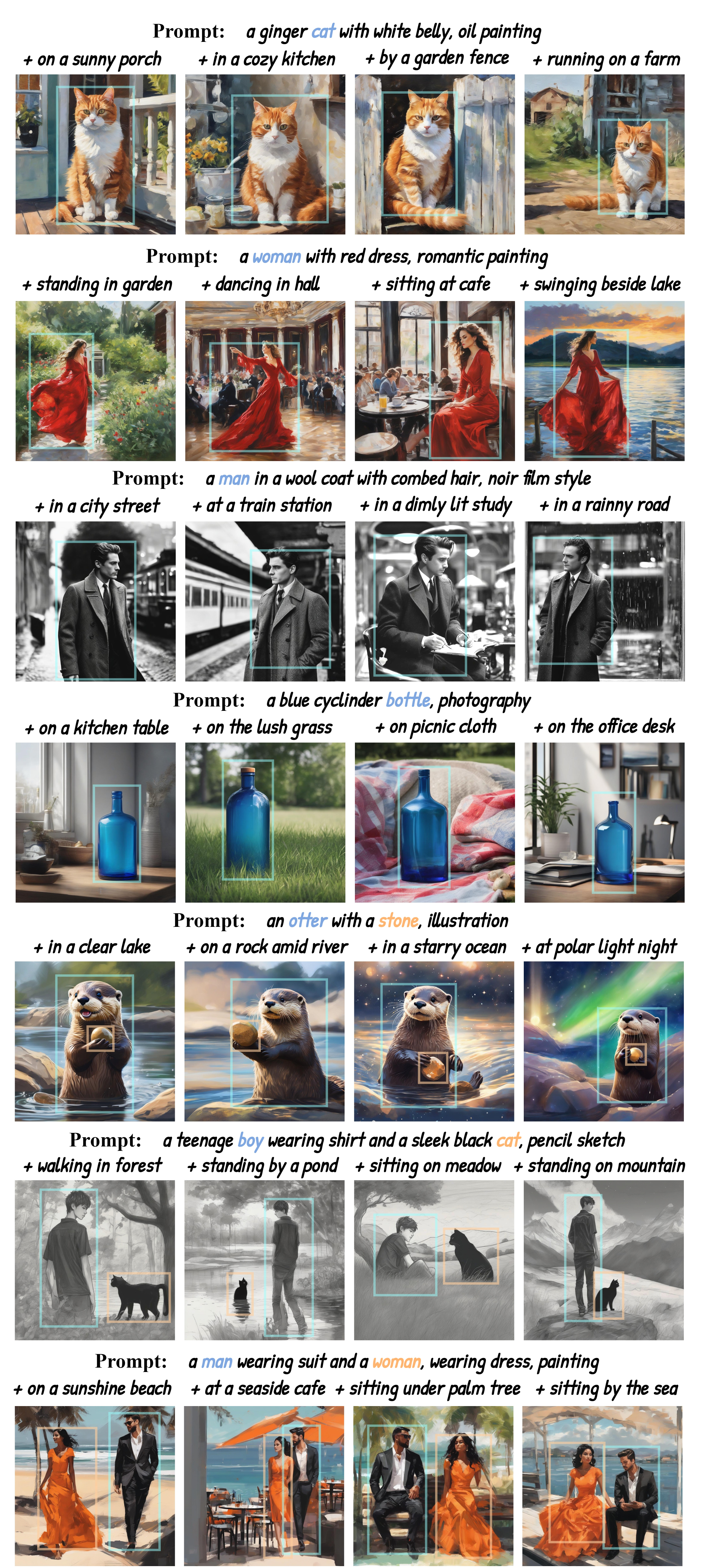}
\caption{More qualitative illustration of SpotActor.}
\label{fig:appendix-vis}
\end{figure}

In order to comprehensively showcase the performance of our method, we provide more generation samples covering single subject and double subjects in Fig.~\ref{fig:appendix-vis}.
\clearpage
\begin{figure*}[t]
\centering
\includegraphics[width=0.9\textwidth]{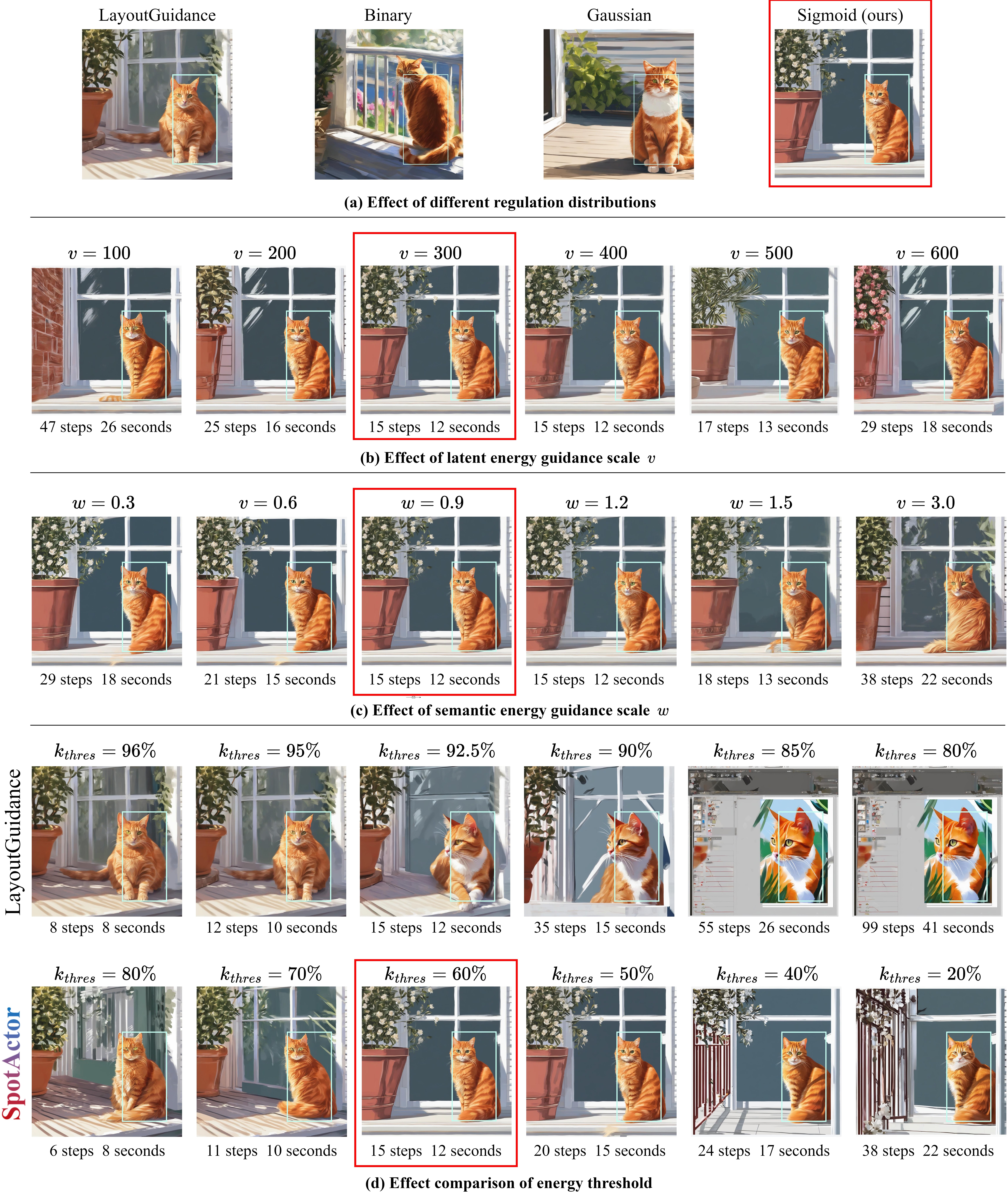}
\caption{Parameter analysis of our method. The best results are outlined in red.}
\label{fig:appendix-para}
\end{figure*}
\clearpage
\newpage
\section{Quantitative Metric Distribution}
\begin{table}[h]
\centering
\setlength{\tabcolsep}{4mm}{
\begin{tabular}{c|ccc}
\hline
Metric      & Aver & 80\% CI   & 95\% CI   \\ \hline
mIoU(↑)     & 67.1 & 63.2-70.4 & 61.4-72.8 \\
DINO-fg(↑)  & 78.6 & 74.4-82.8 & 72.1-83.3 \\
CLIP-fg(↑)  & 79.7 & 75.4-83.1 & 72.6-84.7 \\
LPIPS-fg(↓) & 34.9 & 32.7-35.9 & 30.5-36.8 \\
CLIP-T(↑)   & 63.6 & 61.3-65.0 & 60.9-65.5 \\
DINO-bg(↓)  & 37.8 & 35.1-39.9 & 33.7-41.6 \\
CLIP-bg(↓)  & 49.5 & 45.2-53.8 & 43.8-55.1 \\
LPIPS-bg(↑) & 56.8 & 54.9-58.5 & 54.0-59.6 \\ \hline
\end{tabular}
\caption{Distribution of the metrics. Aver denotes the average value. $n\%$ CI denotes the confidence interval at $n\%$ confidence probability.}
}
\end{table}

\section{Parameter Analysis}
\subsection{Target Function}
The key to layout control in our method is the nuanced layout energy based on Sigmoid-like target functions. To varify the effectiveness of this setting, we construct 3 baseline models: LayoutGuidance, our method using Binary mask as target distribution and our method using Gaussian. We illustrate the comparison of generations in the same setting in Fig.~\ref{fig:appendix-para}(a). It shows that Binary masks and Gaussian function both exceed the given box boundary. By contrast, the Sigmoid function is the best choice which perfectly mimics the inherent activation distribution and aligns the subject to the given box.
\subsection{Latent Energy Guidance Scale}
The latent guidance scale $v$ is the core parameter in energy guidance paradigm. To investigate its influence, we conduct a parameter analysis in Fig.~\ref{fig:appendix-para}(b). As shown, with $v$ increasing, there is an initial reduction in convergence time and image quality, followed by an eventual increase. The observed behavior is attributable to $v$ acting in a manner analogous to the learning rate. When $v$ is too small, the update process slows down, and when too large, oscillations arise, making convergence challenging. We set $v=300$, which is in the optimal interval.
\subsection{Semantic Energy Guidance Scale}
As the other half of the proposed dual energy guidance, we also conduct analysis of $w$, the semantic energy guidance scale in Fig.~\ref{fig:appendix-para}(c). Generally, its effect highly resembles that of $v$, which echos our motivation that the semantic update is inherently entangled with the latent update. We set $w=0.9$ for an optimal performance.
\subsection{Loss Threshold}
Since our method is an update-based approach, the energy threshold $k_\textit{thres}$ of convergence criteria plays a crucial role. Thus, we conduct a comparison on it with LayoutGuidance in Fig.~\ref{fig:appendix-para}(d). It can be observed that LayoutGuidance exhibits a highly constrained adjustable space, with convergence in a critical state, which makes it prone to image degradation. By contrast, our method demonstrates strong adjustability across a wide range, providing evidence of its robust convergence. We set $k_\textit{thres}=60\%$ throughout our paper.

\section{Limitation}
\begin{figure}[h]
\centering
\includegraphics[width=0.45\textwidth]{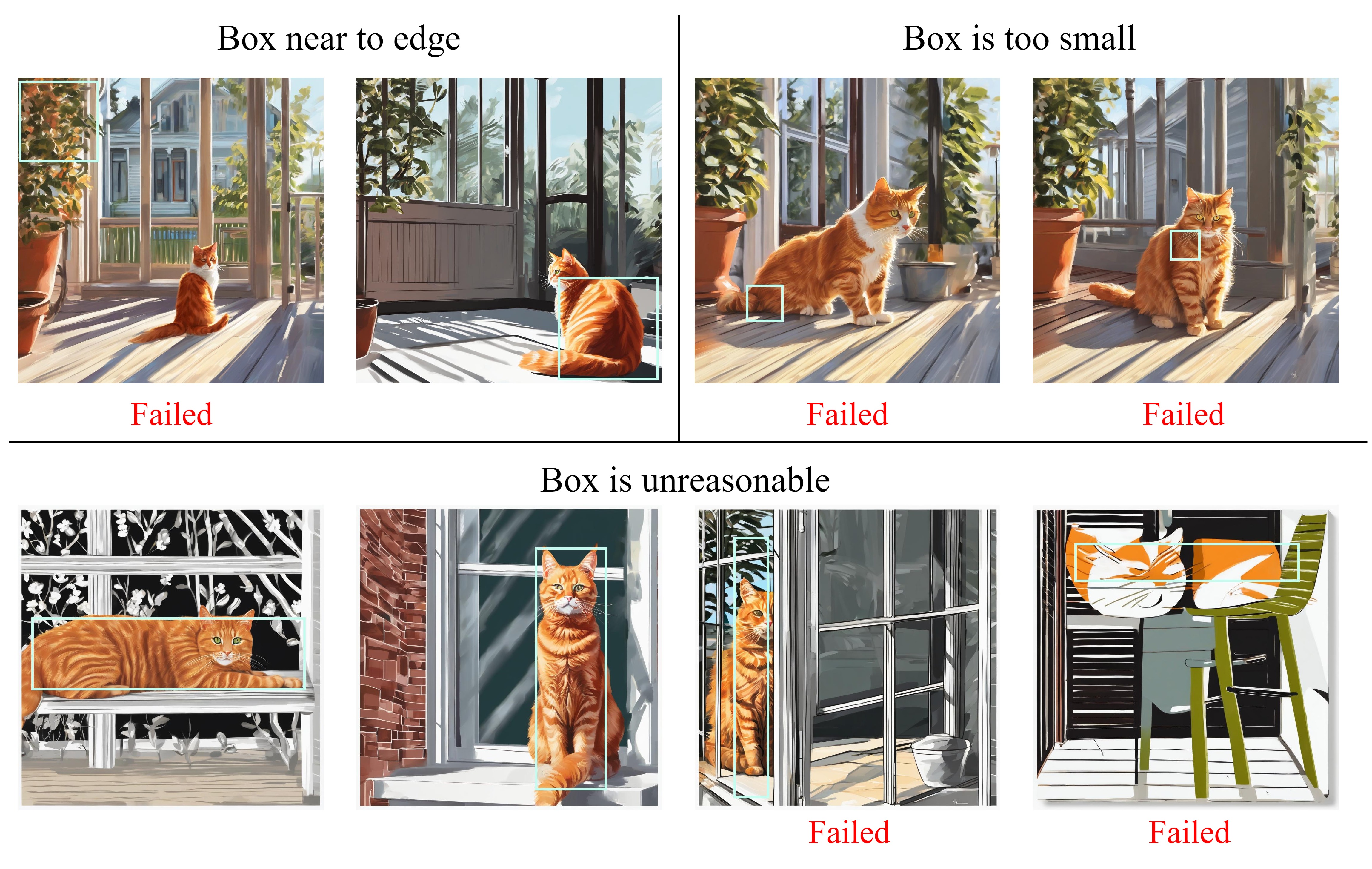}
\caption{Failure cases of our method.}
\label{fig:appendix-lim}
\end{figure}
Despite the superior performance, SpotActor exhibits certain limitations. Based on energy guidance, our method is essentially a score-matching process which finds the optimal sample aligned best to the given layout. Thus, the capacity of our method is highly reliant on the trained distribution of the base model, which results in out-of-distribution (OOD) issues. We show some extreme cases in Fig.~\ref{fig:appendix-lim}. It can be observed that when the given box are too small or near to the edge, our method suffers from poor layout alignment. Because SDXL, after preference alignment, is skilled at generating the main element centering and occupying a significant portion of the image. Meanwhile, failure occasional occurs for boxes that are unreasonable, which stems from SDXL's lack of training data in accordance with such layout. To address this OOD dilemma, incorporating more data remains an effective solution.

\section{Societal Impacts}
The progress in consistent subject generation brings about significant societal impacts across multiple domains. Our work, which enables layout control in text-to-image consistent generation, democratizes the artistic process by enabling creators of all skill levels to produce consistent, high-quality visuals. This innovation streamlines workflows in animation, advertising, and publishing, leading to significant time and cost reductions. The potential to maintain visual consistency in media and entertainment could revolutionize fields such as animation and storybook illustration.

However, the rapid progress of creative diffusion models could lead to job displacement for artists and designers who depend on traditional techniques, resulting in economic challenges and the need for reskilling. Furthermore, ethical issues arise from the potential misuse of generated content, such as the creation of deepfakes or misleading images, which could damage public trust and facilitate misinformation. Additionally, ensuring that generative models are trained on unbiased datasets is vital to avoid perpetuating harmful stereotypes or biases. To address these concerns, our model, based on Stable Diffusion, incorporates safeguards like NSFW detection and adherence to usage guidelines.

\end{document}